\documentclass[11pt]{article}
\usepackage{geometry}
\usepackage{authblk}
\usepackage{graphicx}
\usepackage{coling2020}
\usepackage{times}
\usepackage[hyphens]{url}
\usepackage{latexsym}
\usepackage{microtype}
\usepackage{makecell}
\usepackage{latexsym}
\usepackage{tabularx,booktabs}
\usepackage{calrsfs}
\usepackage{fixltx2e}
\usepackage{xcolor}
\usepackage[multiple]{footmisc}

\DeclareMathAlphabet{\pazocal}{OMS}{zplm}{m}{n}
\newcommand{\Lb}{\pazocal{L}}
\newcolumntype{Y}{>{\centering\arraybackslash}X}

\newcommand{\comment}[1]{}

\colingfinalcopy 

\title{UPB at SemEval-2020 Task 8: Joint Textual and Visual Modeling in a Multi-Task Learning Architecture for Memotion Analysis}

\author{George-Alexandru Vlad, George-Eduard Zaharia, Dumitru-Clementin Cercel, \\
 Costin-Gabriel Chiru, Stefan Trausan-Matu\\
  University Politehnica of Bucharest, Faculty of Automatic Control and Computers \\
  {\tt \{george.vlad0108,george.zaharia0806\}@stud.acs.upb.ro} \\
  \tt\{dumitru.cercel,costin.chiru,stefan.trausan\}@cs.pub.ro}

\begin{document}
\maketitle
\begin{abstract}
 Users from the online environment can create different ways of expressing their thoughts, opinions, or conception of amusement. Internet memes were created specifically for these situations. Their main purpose is to transmit ideas by using combinations of images and texts such that they will create a certain state for the receptor, depending on the message the meme has to send. These posts can be related to various situations or events, thus adding a funny side to any circumstance our world is situated in. In this paper, we describe the system developed by our team for SemEval-2020 Task 8: Memotion Analysis. More specifically, we introduce a novel system to analyze these posts, a multimodal multi-task learning architecture that combines ALBERT for text encoding with VGG-16 for image representation. In this manner, we show that the information behind them can be properly revealed. Our approach achieves good performance on each of the three subtasks of the current competition, ranking  \(11^{th}\) for Subtask A (0.3453 macro F1-score), \(1^{st}\) for Subtask B (0.5183 macro F1-score), and \(3^{rd}\) for Subtask C (0.3171 macro F1-score) while exceeding the official baseline results by high margins.
\end{abstract}

\section{Introduction}
\label{intro}
\blfootnote{This work is licensed under a Creative Commons Attribution 4.0 International Licence. \\ \hspace*{1.6em} Licence details: http://creativecommons.org/licenses/by/4.0/.
}
The Internet represents the biggest source of knowledge humankind currently possesses. People from all corners of the world can express their opinions, thoughts, and share their insights regarding certain ideas. During the past decade, a new and never seen before way of sharing beliefs arose, memes. For example, by humorously combining text and images, their authors can emphasize a series of aspects such that they will amuse, in various degrees and ways, the receptors.

\textbf{\textit{Motivation}}. Internet memes come with various templates and formats. Some of them are purely humorous, while others, behind an amusing appearance, intend to convey subtle nuances including sarcasm, disbelief regarding an idea, or a motivational purpose. 
All of them are present in the online environment and offer 
an insight into the opinion of some communities regarding particular aspects. Moreover, they can be used for obtaining valuable information that will lead to further improvements for the web content mining process.

\textbf{\textit{Challenges}}. The mining task becomes increasingly more difficult, since
both the image and text clarity usually seem to vary substantially, depending on the user or the region it was posted from. 
These situations may cause unfavorable results when performing an analysis process and, in particular cases, they can transmit a different idea than intended. If the ambiguity reaches high values, it can lead to the impossibility of separating the two main ways memes convey information: text and image. Also, since the text is embedded into the image, a low-quality picture can introduce noise inside the content and can compromise the entire meme. On the other way around, an extremely unclear text will create a discrepancy between it and the visual aspect of the post, and thus can weaken the overall message. 

\comment{
Figure \ref{fig:pic1} shows an example of ambiguous and well-created memes\footnote{\url{https://ahseeit.com/tamil/king-include/uploads/2019/04/54732127_161097838179106_8464561624281184655_n-332137327.jpg}}\footnote{\url{https://amazinganimalphotos.com/wp-content/uploads/2019/02/cat-memes-2019-1.jpg}}.

}

The  memotion analysis shared task  \cite{chhavi2020memotion} organized by SemEval-2020 intends to challenge participants to approach the previously mentioned issues to create systems able to analyze Internet memes. The competition consists of three subtasks. Subtask A intends to properly classify the memes as either positive, neutral, or negative. Furthermore, Subtask B builds upon the first subtask such that the participants will be challenged to binary classify the posts considering four categories: humor, sarcasm, offense, and motivation. Finally, Subtask B is extended into Subtask C, where the four categories are expanded into  classes of different granularity, gradually increasing from the lowest to the highest possibility regarding to each category.

\textbf{\textit{Proposed Approach}}. We intend to solve all the previously mentioned subtasks by introducing a neural network based on multi-task learning (MTL). The system will contain modules dedicated to image analysis and modules specialized in text processing. The architecture outputs a single answer for all the required subtasks.

\comment{
\begin{figure}[t!]
\centering
\includegraphics[width=0.7\linewidth]{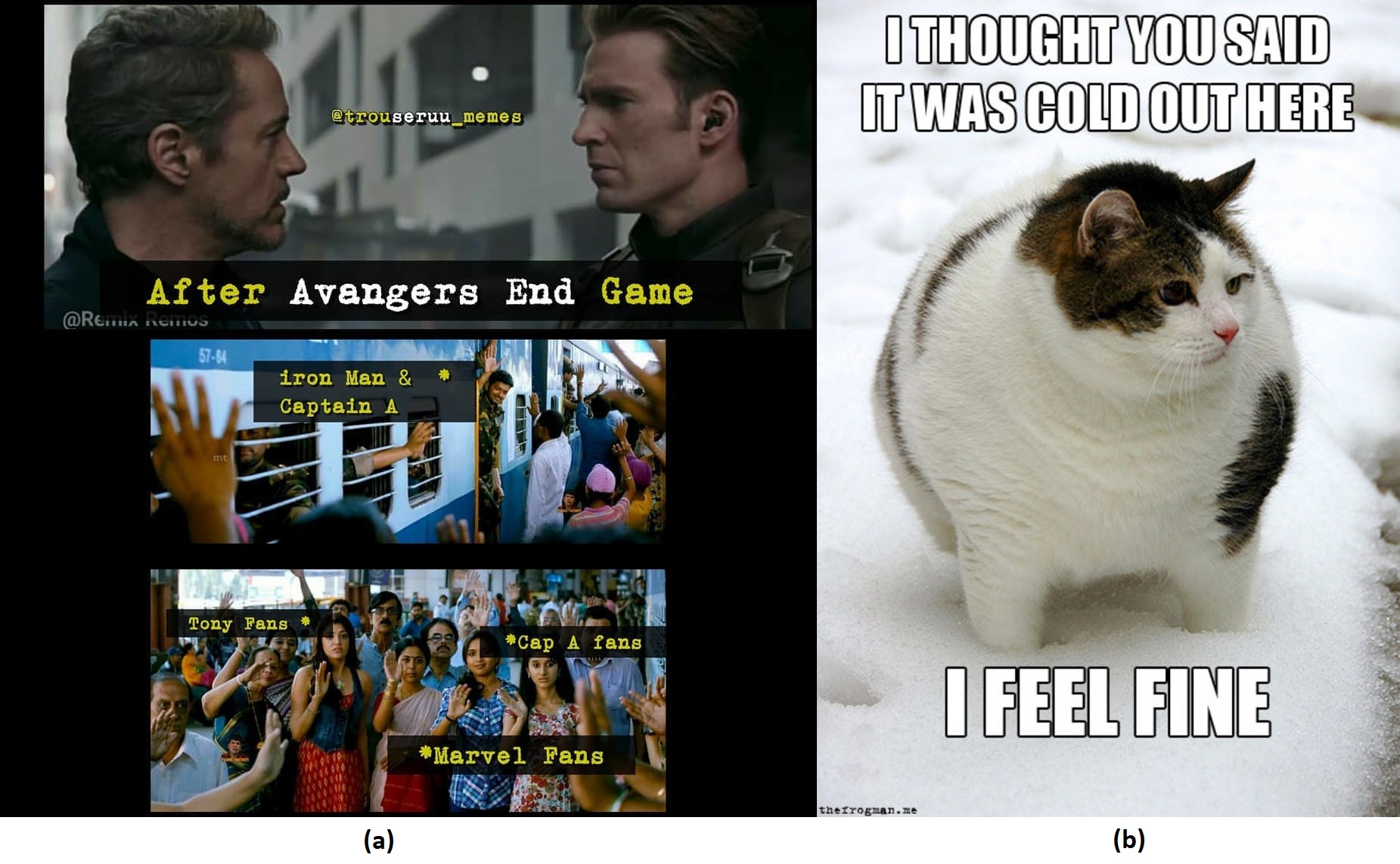}
  \caption{(a) Example of an ambiguous meme: a combination of different images, alongside poorly written text overlapping another text and (b) Good meme with high-quality image and straightforward text.}
  \label{fig:pic1}
\end{figure}

}

The next parts of this work are structured as follows. In section 2, we perform an analysis of  existing solutions found on  related works. In section 3, we outline the approaches we applied for memotion analysis. Section 4 details the performed experiments,  experimental setup, and error analysis. Finally, we draw conclusions in section 5.

\section{Related Work}

\subsection{Humor Recognition}

\newcite{yoshida:2018} proposed a multimodal approach regarding humor identification, by combining a Convolutional Neural Network (CNN)~\cite{fukushima:neocognitronbc} and a Long Short-term Memory Network~\cite{lstm_hochreiter}. The authors also used a ResNet-152~\cite{he2015deep} model and developed a custom loss function that takes into account a \textit{funny score}.  To compute the \textit{funniness} of a post, they used the reviews from a certain humor website, where people use stars to evaluate them. The loss function revolves around a certain threshold, set by the authors to be 100.

Also, humor recognition has been addressed by using a feature-based solution \cite{chandrasekaran:2016}. The authors created features by analyzing the inputs on different levels: cardinality, location, object, instance-level features. A Support Vector Regression  is used for humor score prediction. Furthermore, the authors introduced a new technique for improving the humor scores, by altering the \textit{funniness} of a scene. Firstly, they detected the objects that contribute to the humor of that particular scene, and then they identified replacements that can alter the \textit{funniness} of the scene. For the former part, they used a multi-layer perceptron (MLP) that outputs a binary class for each object in the scene. Then, for the latter part, altering the humor, they used another MLP trained for identifying potential replacements for the original object.

\subsection{Multimodal Classification}
 Besides humor identification, sentiment analysis is another process that can be applied to image-text pair entries \cite{qian:2019}.  The authors introduced an architecture based on a CNN inspired by AlexNet~\cite{krizhevsky:2012}, as well as a combination of Support Vector Machines and AffectiveSpace 2~\cite{cambria:2015}. Regarding the visual features, a difference from AlexNet is represented by the replacements of multiple fully connected layers with a single fully connected layer of size 4096 x 2. Furthermore, the textual features are extracted by using a 5-fold cross-validation based on AffectiveSpace 2. Their system surpasses the other existing solutions by a margin of 7.2\%. The authors concluded that the combination of visual and textual features offers improved results when compared to a single modality.

EmbraceNet \cite{choi:2019} intends to improve the results of multimodal classification tasks by offering flexibility for any learning model, as well as various methods of dealing with absent data. Their architecture consists of \textit{Docking layers}, used for transforming each input vector to a certain size. Furthermore, the \textit{Embracement layer} is used for combining feature vectors into a single vector, by using a multinomial sampling method. The model also considered the correlation between modalities, handled missing data, and performed a regularization effect. The final idea is that EmbraceNet obtained considerably improved scores compared to its counterparts, representing a good choice for multimodal classification tasks.

\section{Proposed Approaches}
We propose three multi-task learning architectures to solve all the subtasks of the memotion analysis competition (i.e. sentiment classification, humor classification, and scales of semantic classes). Because the entries are bimodal, the resulting system can be divided into two main components based on their input (i.e. one for image-only and the other for text-only). Both parts will act as a feature extractor, the resulting encodings being fused together to obtain the text-image representation. Based on the fused features, our system will discriminate between the output classes of each subtask.
Next, we offer details regarding the three systems.

\subsection{Text-only Multi-task Architecture}
In order to extract the most salient features from the text input, we opted to use the ALBERT model \cite{lan2019albert}, pre-trained on an English corpus. ALBERT obtained state-of-the-art results on the GLUE~\cite{wang2018glue}, RACE~\cite{lai-etal-2017-race}, and SQuAD~\cite{rajpurkar2018know} benchmarks, while being more memory efficient and requiring lower training times than its predecessor, BERT~\cite{devlin2018bert}. Moreover, ALBERT removes the Next Sentence Prediction pre-training strategy and replaces it with Sentence Order Prediction which better ensures the inter-sentence coherence. As a consequence of cross-layer parameter sharing used to decrease memory complexity, the model converges faster and smoother than BERT, proving to be a more stable architecture. The aforementioned improvements of the ALBERT model over BERT represent the main reasons for choosing ALBERT as our text encoding.

The ALBERT model comes in four variants based on the number of its parameters: base (12M), large (18M), xlarge (60M), and xxlarge (235M). We opted for the xlarge variant in order to leverage the trade-off between model size and performance. From an architectural standpoint, ALBERT xlarge has 24 layers with 16 attention heads, each one having a hidden dimension of 2,048 neurons. We extracted the pooled output of ALBERT as the feature vector encoding of the textual input. We added a dropout layer \cite{srivastava2014dropout} of 0.1 over the resulted feature vector as a regularization mechanism in order to ensure the robustness of our model. At this step, the output from the previous layer will be fed to five independent classifiers responsible for each category tracked in the competition. All the classifiers consist of two fully connected layers of size 512 and 256, respectively, each followed by a dropout layer of 0.3. For each subtask, a fully connected layer for the output is added  on top of each such layer-stack, where every neuron is corresponding to a class.
In the end, we obtained five outputs accounting for sentiment classification (three classes), humor classification (four classes), sarcasm classification (four classes), offense classification (four classes), and motivational classification (two classes). We used the softmax activation function on the output layers in order to obtain the distribution probability over the classes.

\subsection{Image-only Multi-task Architecture}
To extract the features from the image input, we opted for the VGG-16 architecture \cite{simonyan2014very}, which consists of five stacks of convolutional layers using a 3x3 kernel size. Furthermore, each stack is followed by max-pooling layers of 2x2 dimension, the last one being linked to three fully connected layers, where the first two have 4,096 units and the third only 1,000 units. We removed the last three layers from the original architecture to obtain only the stacks of convolutions used in extracting the features from the input image. Moreover, we connected the last layer to a global average pooling layer to form a 1-D feature vector. As in the text-only feature extractor model, we connected the obtained feature vector to the independent classifier stack of layers. At the same time, the resulting architecture adopts a transfer learning strategy by using the pre-trained weights on the ImageNet dataset~\cite{deng:2009}. 

In our case, the meme templates vary extensively from a single image with associated text to multiple image crops stacked together. By resizing the images to the input size required by the original VGG-16 architecture (i.e. 224x224 resolution), relevant aspects from the memes might be lost. As a measure to prevent detrimental information loss, we replace the default input layer of the VGG-16 to accept the new images, resizing them to a resolution of 500x500 pixels.

\subsection{Multimodal Multi-task Architecture}
\begin{figure}[h!]
\centering
\includegraphics[width=0.9\linewidth]{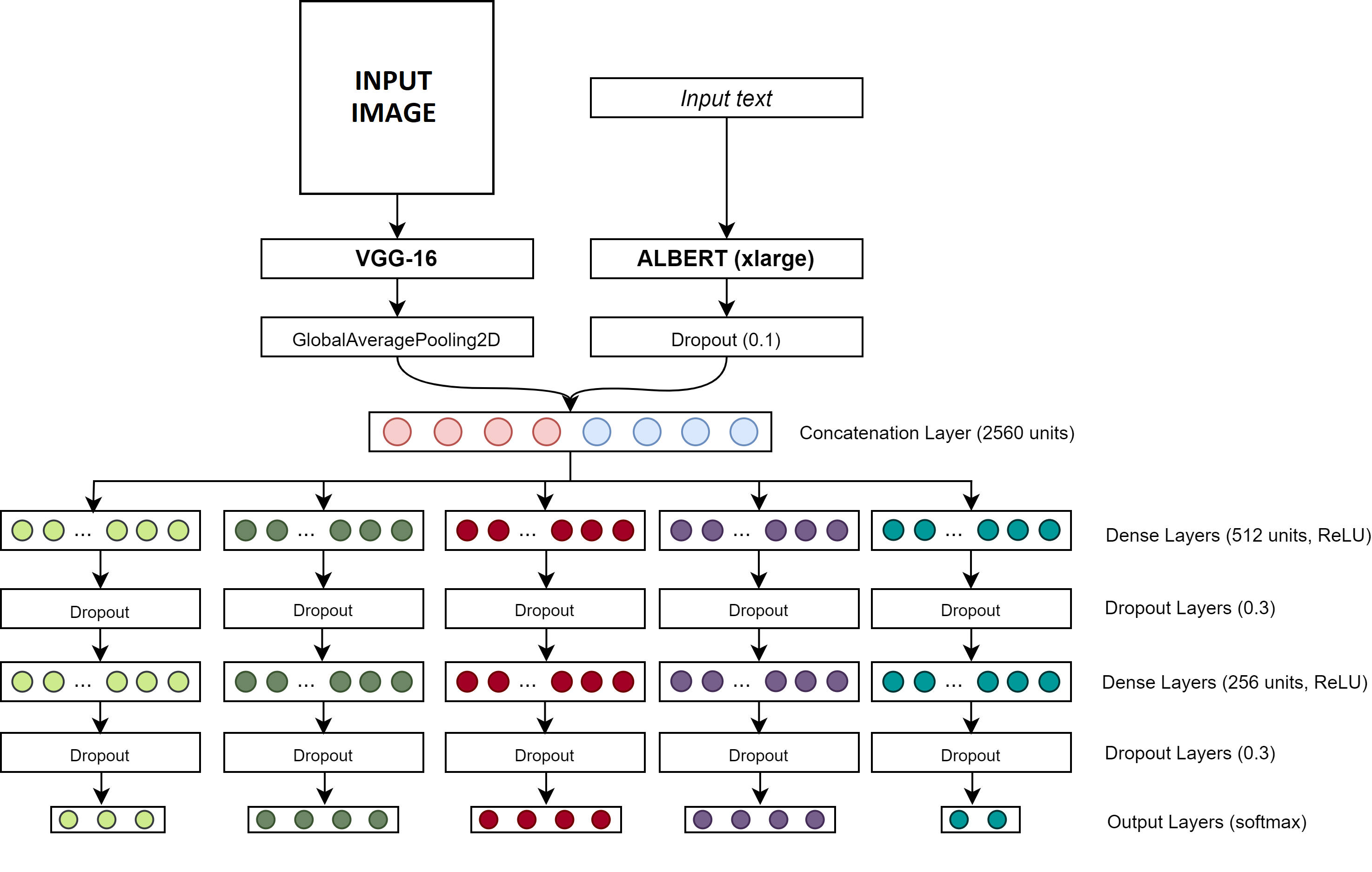}
  \caption{Proposed multimodal architecture with multi-task learning.\comment{\footnotemark}}
  \label{fig:pic2}
\end{figure}

In order to take advantage of the image-text relationship, we combined the two separate textual and visual feature extraction components into a unified architecture. It receives both the input image and the processed text into the ALBERT input format, each input channel being processed independently by the corresponding specialized component. Passing the information through both models results into two feature embeddings \(E_i \in R^{512}\) and \(E_t \in R^{2048}\), for the image and text embeddings, respectively. The resulting embeddings are then concatenated to obtain the image-text vector representation \(E_{it} \in R^{2560}\). This output is sent to the set of independent classifiers for each subtask. In this case, an overview of our architecture is illustrated in Figure \ref{fig:pic2}.

\comment{
\footnotetext{\url{https://66.media.tumblr.com/541912accdc7359450327f825821098a/tumblr_otl9vaOwVa1wnk95eo1_1280.jpg}}
}

\section{Experiments}
\subsection{Data and Preprocessing}
The dataset consists of 6,992 memes for training, out of which we randomly selected 10\% for validation, and 2,000 memes for testing, provided in different image formats. Alongside the memes, an extra file specifies the labels (i.e. humor, sarcasm, offensive, motivational, and overall sentiment), as well as the extracted text, for each one of them. The class distribution is imbalanced, thus implying the usage of loss weights in the loss function \cite{khosla2018emotionx}.

The text is preprocessed such that it will become a proper input for the ALBERT model  as follows: tokenization is done by using the SentencePiece subword tokenizer \cite{kudo2018sentencepiece} officially released with ALBERT, followed by the computation of the \textit{input ids} (i.e., token’s position in the vocabulary file associated with SentencePiece), the \textit{input masks} (namely, values of 1 for the actual tokens to be considered by the model, or values of 0 for the padding tokens), and the \textit{segment ids} (representing the sentence to which the input tokens belong to). As previously mentioned, the images contained in the dataset are resized to a convenient size of 500x500, and then are serialized to \textit{tfrecord} files alongside the corresponding text and labels. For the test set, the preprocessing pipeline is similar to the one for training.

\subsection{Experimental Setup}
We conducted three experiments, one for each input-specific MTL architecture.
Each of them has been trained in two steps. In the first step, we opted to freeze all the layers for the feature extracting component (i.e. ALBERT or VGG-16), depending on the input type. During the second step, we unfroze the weights and allowed the networks to properly adjust them.
The architectures have been implemented using TensorFlow 2.1~\cite{abadi2016tensorflow}. Furthermore, we experimented with two different optimizers, Adam~\cite{kingma2014adam} and LAMB~\cite{you2019large}, in order to maximize the performance of our solutions. We also introduced a patience of 30, such that our models will stop training if no improvements have occurred during the last 30 epochs. Because we trained all our architectures in two steps, during the first one, we set a learning rate of \textit{5e-4} with a warm-up over the first 10\% of the total training steps. The second step ran with a peak learning rate of \textit{5e-5}. Table \ref{table:table1} presents the hyper-parameters used for training the models during the experiments.

\begin{table}[h]

\begin{center}
\caption{\label{table:table1} Neural model hyper-parameters. }
\begin{tabular}{|l|l|}
\hline \bf Hyper-parameter & \bf Values  \\ \hline
\textbf{optimizer} & LAMB, AdamWeightDecay \\
\textbf{learning rate} & frozen: \textit{5e-4}, unfrozen: \textit{5e-5} \\
\textbf{patience} & 30  \\
\textbf{classifier dropout} & 0.1  \\
\textbf{warmup proportion} & 0.1 \\
\textbf{weight decay} & 0.01 \\
\textbf{Adam epsilon} & \textit{1e-6} \\
\hline
\end{tabular}
\end{center}

\end{table}

Furthermore, we used five separate loss functions, namely $ \Lb $\textsubscript{sentiment}, $ \Lb $\textsubscript{humor}, $ \Lb $\textsubscript{sarcasm}, $ \Lb $\textsubscript{offense}, and $ \Lb $\textsubscript{motivation}, where each one represents a categorical crossentropy and has the following format:

\[  \Lb_{x} = -\frac{1}{N}\sum_{i=1}^{N}log~p_{model}~[y_{i} \in C_{y_{i}}]  \] 

The term y\textsubscript{i} represents the observation, while C is the output class.

\subsection{Results}

\begin{table}[h!]
\centering
\caption{Experimental results (macro F1-score) of our architectures on the validation set. In our case, considering the competition's evaluation system, the results for Subtask C can be used for determining the results for Subtask B, inasmuch as the latter lowers the specificity for the labels used for the former.}
\label{table:table2}
\resizebox{\textwidth}{!}{\begin{tabular}{|c|c|c|c|c|c|}
\hline

\bf \makecell{Architecture} & \bf\makecell{Subtask A \\ (Sentiment)} & \bf \makecell{Subtask C \\ (Humor)} & \bf \makecell{Subtask C \\ (Sarcasm)} &  \bf \makecell{Subtask C \\ (Offense)} & \bf \makecell{Subtask C \\ (Motivation)}  \\ 

\hline

VGG-16 only MTL & 0.3355 &  0.2762 &  0.2836 & 0.2693 & 0.4983 \\
ALBERT only MTL & 0.4399 &  0.4375 &  0.4391 & \bf 0.4255 & 0.5652 \\
Multimodal Fusion MTL & \bf 0.4415 &  \bf 0.4436 &  \bf 0.4511 & 0.3875 & \bf 0.5829 \\
\hline

\end{tabular}}
\end{table}

Table \ref{table:table2} contains the results obtained from running the experiments. As expected, the best results are yielded by the usage of the multimodal method, the MTL ensemble comprised of ALBERT for text encoding, and VGG-16 for image processing. Moreover, the text-only MTL solution is very close to the best results obtained by the multimodal MTL. For example, for the sentiment identification subtask, the multimodal MTL solution manages to achieve a macro F1-score just 0.16\% higher compared to the ALBERT-only MTL, even though the number of parameters is considerably larger when compared to the text-only counterpart.
This result might occur due to the lack of information contained in most of the meme images, often the text being the crucial factor in establishing the final classification. This can also be found when comparing the results obtained by using only the image part of the network, VGG-16. At the same time, there is a difference of 15.62\% macro F1-score for the offense identification subtask between the VGG-16 MTL network and the ALBERT  MTL solution. This difference is further justified by the fact that the text produces a larger number of features, in contrast to images. Furthermore, on the test set, we obtained the following results, specified in the official competition format (i.e. three main subtasks): 0.3453 macro F1-score for Subtask A, 0.5183 macro F1-score for Subtask B, and 0.3171 macro F1-score for Subtask C, respectively.

\section{Conclusions and Future Works}
This paper presented our approaches regarding the memotion analysis shared task, organized by SemEval-2020.
We proposed several architectures that intends to solve the memotion analysis issue by using recent breakthroughs in the computer vision, as well as the natural language processing field: VGG-16 alongside ALBERT.
By creating multi-task learning solutions, joint textual and visual modeling network, and separate channels, as well, we were able to achieve good scores for the previously mentioned subtasks and provide a good insight on the way this relatively new challenge can be approached.
We highlighted that factors such as image noise, incorrect typing, or overlapping text are key parts that stand against a proper analysis.
Also, various combinations of texts and images can cause serious problems in the analysis process conducted by deep learning approaches.

In the future work, we will investigate the effect of using improved visual models, such as the enhancement for VGG-16, VGG-19~\cite{simonyan2014very}. Furthermore, considering that ALBERT is a lite BERT, we also intend to explore text processing models containing a larger number of parameters, such as BERT-base, or even BERT-large.

\section*{Acknowledgements}
The work has been funded by the Operational Programme Human Capital of the Ministry of European Funds through the Financial Agreement
51675/09.07.2019, SMIS code 125125.

\bibliographystyle{coling}
\bibliography{semeval2020}

\begin{thebibliography}{}

\bibitem[\protect\citename{Abadi \bgroup et al.\egroup
  }2016]{abadi2016tensorflow}
Mart{\'\i}n Abadi, Ashish Agarwal, Paul Barham, Eugene Brevdo, Zhifeng Chen,
  Craig Citro, Greg~S Corrado, Andy Davis, Jeffrey Dean, Matthieu Devin, et~al.
\newblock 2016.
\newblock Tensorflow: Large-scale machine learning on heterogeneous distributed
  systems.
\newblock {\em arXiv preprint arXiv:1603.04467}.

\bibitem[\protect\citename{Cambria \bgroup et al.\egroup }2015]{cambria:2015}
Erik Cambria, Jie Fu, Federica Bisio, and Soujanya Poria.
\newblock 2015.
\newblock Affectivespace 2: Enabling affective intuition for concept-level
  sentiment analysis.
\newblock In {\em Twenty-ninth AAAI conference on artificial intelligence}.

\bibitem[\protect\citename{Chandrasekaran \bgroup et al.\egroup
  }2016]{chandrasekaran:2016}
Arjun Chandrasekaran, Ashwin~K Vijayakumar, Stanislaw Antol, Mohit Bansal,
  Dhruv Batra, C~Lawrence~Zitnick, and Devi Parikh.
\newblock 2016.
\newblock We are humor beings: Understanding and predicting visual humor.
\newblock In {\em Proceedings of the IEEE Conference on Computer Vision and
  Pattern Recognition}, pages 4603--4612.

\bibitem[\protect\citename{Choi and Lee}2019]{choi:2019}
Jun-Ho Choi and Jong-Seok Lee.
\newblock 2019.
\newblock Embracenet: A robust deep learning architecture for multimodal
  classification.
\newblock {\em Information Fusion}, 51:259--270.

\bibitem[\protect\citename{Deng \bgroup et al.\egroup }2009]{deng:2009}
Jia Deng, Wei Dong, Richard Socher, Li-Jia Li, Kai Li, and Li~Fei-Fei.
\newblock 2009.
\newblock Imagenet: A large-scale hierarchical image database.
\newblock In {\em 2009 IEEE conference on computer vision and pattern
  recognition}, pages 248--255.

\bibitem[\protect\citename{Devlin \bgroup et al.\egroup }2019]{devlin2018bert}
Jacob Devlin, Ming-Wei Chang, Kenton Lee, and Kristina Toutanova.
\newblock 2019.
\newblock Bert: Pre-training of deep bidirectional transformers for language
  understanding.
\newblock In {\em Proceedings of the 2019 Conference of the North American
  Chapter of the Association for Computational Linguistics: Human Language
  Technologies, Volume 1 (Long and Short Papers)}, pages 4171--4186.

\bibitem[\protect\citename{Fukushima and Miyake}1982]{fukushima:neocognitronbc}
Kunihiko Fukushima and Sei Miyake.
\newblock 1982.
\newblock Neocognitron: A self-organizing neural network model for a mechanism
  of visual pattern recognition.
\newblock In {\em Competition and cooperation in neural nets}, pages 267--285.
  Springer.

\bibitem[\protect\citename{He \bgroup et al.\egroup }2016]{he2015deep}
Kaiming He, Xiangyu Zhang, Shaoqing Ren, and Jian Sun.
\newblock 2016.
\newblock Deep residual learning for image recognition.
\newblock In {\em Proceedings of the IEEE conference on computer vision and
  pattern recognition}, pages 770--778.

\bibitem[\protect\citename{Hochreiter and Schmidhuber}1997]{lstm_hochreiter}
Sepp Hochreiter and J{\"u}rgen Schmidhuber.
\newblock 1997.
\newblock Long short-term memory.
\newblock {\em Neural computation}, 9(8):1735--1780.

\bibitem[\protect\citename{Khosla}2018]{khosla2018emotionx}
Sopan Khosla.
\newblock 2018.
\newblock Emotionx-ar: Cnn-dcnn autoencoder based emotion classifier.
\newblock In {\em Proceedings of the Sixth International Workshop on Natural
  Language Processing for Social Media}, pages 37--44.

\bibitem[\protect\citename{Kingma and Ba}2014]{kingma2014adam}
Diederik~P Kingma and Jimmy Ba.
\newblock 2014.
\newblock Adam: A method for stochastic optimization.
\newblock {\em arXiv preprint arXiv:1412.6980}.

\bibitem[\protect\citename{Krizhevsky \bgroup et al.\egroup
  }2012]{krizhevsky:2012}
Alex Krizhevsky, Ilya Sutskever, and Geoffrey~E Hinton.
\newblock 2012.
\newblock Imagenet classification with deep convolutional neural networks.
\newblock In {\em Advances in neural information processing systems}, pages
  1097--1105.

\bibitem[\protect\citename{Kudo and Richardson}2018]{kudo2018sentencepiece}
Taku Kudo and John Richardson.
\newblock 2018.
\newblock Sentencepiece: A simple and language independent subword tokenizer
  and detokenizer for neural text processing.
\newblock In {\em Proceedings of the 2018 Conference on Empirical Methods in
  Natural Language Processing: System Demonstrations}, pages 66--71.

\bibitem[\protect\citename{Lai \bgroup et al.\egroup }2017]{lai-etal-2017-race}
Guokun Lai, Qizhe Xie, Hanxiao Liu, Yiming Yang, and Eduard Hovy.
\newblock 2017.
\newblock Race: Large-scale reading comprehension dataset from examinations.
\newblock In {\em Proceedings of the 2017 Conference on Empirical Methods in
  Natural Language Processing}, pages 785--794.

\bibitem[\protect\citename{Lan \bgroup et al.\egroup }2019]{lan2019albert}
Zhenzhong Lan, Mingda Chen, Sebastian Goodman, Kevin Gimpel, Piyush Sharma, and
  Radu Soricut.
\newblock 2019.
\newblock Albert: A lite bert for self-supervised learning of language
  representations.
\newblock {\em arXiv preprint arXiv:1909.11942}.

\bibitem[\protect\citename{Qian \bgroup et al.\egroup }2019]{qian:2019}
Chen Qian, Edoardo Ragusa, Iti Chaturvedi, Erik Cambria, and Rodolfo Zunino.
\newblock 2019.
\newblock Text-image sentiment analysis.

\bibitem[\protect\citename{Rajpurkar \bgroup et al.\egroup
  }2018]{rajpurkar2018know}
Pranav Rajpurkar, Robin Jia, and Percy Liang.
\newblock 2018.
\newblock Know what you don’t know: Unanswerable questions for squad.
\newblock In {\em Proceedings of the 56th Annual Meeting of the Association for
  Computational Linguistics (Volume 2: Short Papers)}, pages 784--789.

\bibitem[\protect\citename{Sharma \bgroup et al.\egroup
  }2020]{chhavi2020memotion}
Chhavi Sharma, Deepesh Bhageria, William Paka, Scott, Srinivas P~Y K~L, Amitava
  Das, Tanmoy Chakraborty, Viswanath Pulabaigari, and Bj{\"o}rn Gamb{\"a}ck.
\newblock 2020.
\newblock {SemEval-2020 Task 8: Memotion Analysis-The Visuo-Lingual Metaphor!}
\newblock In {\em Proceedings of the 14th International Workshop on Semantic
  Evaluation ({S}em{E}val-2020)}, Barcelona, Spain, Sep. Association for
  Computational Linguistics.

\bibitem[\protect\citename{Simonyan and Zisserman}2014]{simonyan2014very}
Karen Simonyan and Andrew Zisserman.
\newblock 2014.
\newblock Very deep convolutional networks for large-scale image recognition.
\newblock {\em arXiv preprint arXiv:1409.1556}.

\bibitem[\protect\citename{Srivastava \bgroup et al.\egroup
  }2014]{srivastava2014dropout}
Nitish Srivastava, Geoffrey Hinton, Alex Krizhevsky, Ilya Sutskever, and Ruslan
  Salakhutdinov.
\newblock 2014.
\newblock Dropout: a simple way to prevent neural networks from overfitting.
\newblock {\em The journal of machine learning research}, 15(1):1929--1958.

\bibitem[\protect\citename{Wang \bgroup et al.\egroup }2018]{wang2018glue}
Alex Wang, Amanpreet Singh, Julian Michael, Felix Hill, Omer Levy, and Samuel
  Bowman.
\newblock 2018.
\newblock Glue: A multi-task benchmark and analysis platform for natural
  language understanding.
\newblock In {\em Proceedings of the 2018 EMNLP Workshop BlackboxNLP: Analyzing
  and Interpreting Neural Networks for NLP}, pages 353--355.

\bibitem[\protect\citename{Yoshida \bgroup et al.\egroup }2018]{yoshida:2018}
Kota Yoshida, Munetaka Minoguchi, Kenichiro Wani, Akio Nakamura, and Hirokatsu
  Kataoka.
\newblock 2018.
\newblock Neural joking machine: Humorous image captioning.
\newblock {\em arXiv preprint arXiv:1805.11850}.

\bibitem[\protect\citename{You \bgroup et al.\egroup }2019]{you2019large}
Yang You, Jing Li, Sashank Reddi, Jonathan Hseu, Sanjiv Kumar, Srinadh
  Bhojanapalli, Xiaodan Song, James Demmel, Kurt Keutzer, and Cho-Jui Hsieh.
\newblock 2019.
\newblock Large batch optimization for deep learning: Training bert in 76
  minutes.
\newblock {\em arXiv preprint arXiv:1904.00962}.

\end{thebibliography}

\end{document}